\documentclass{article}

\usepackage{microtype}
\usepackage{graphicx}
\usepackage{svg}
\usepackage{enumitem}

\usepackage{booktabs} %

\usepackage{hyperref}
\usepackage[accepted]{icml2019}

\usepackage{enumitem}
\usepackage{amsfonts}       %
\usepackage{makecell}
\usepackage{dcolumn}
\usepackage{caption}
\usepackage{subcaption}
\usepackage{tikz}

\usetikzlibrary{tikzmark, positioning, arrows, decorations.pathreplacing, calc}

\newcommand{\ourmodel}[0]{{MixHop}}
\usepackage{amsmath}

\usepackage{arydshln}
\makeatletter
\renewcommand*\env@matrix[1][*\c@MaxMatrixCols c]{%
	\hskip -\arraycolsep
	\let\@ifnextchar\new@ifnextchar
	\array{#1}}
\makeatother

\newtheorem{definition}{Definition}
\newtheorem{theorem}{Theorem}
\usepackage{amssymb}

\icmltitlerunning{\ourmodel: Higher-Order Graph Convolution Architectures via Sparsified Neighborhood Mixing}

\begin{document}

\twocolumn[
\icmltitle{\ourmodel: Higher-Order Graph Convolutional Architectures \\ via Sparsified Neighborhood Mixing}

\icmlsetsymbol{equal}{*}

\begin{icmlauthorlist}
\icmlauthor{Sami Abu-El-Haija}{isi}
\icmlauthor{Bryan Perozzi}{goog}
\icmlauthor{Amol Kapoor}{goog}
\icmlauthor{Nazanin Alipourfard}{isi}
\icmlauthor{Kristina Lerman}{isi}
\icmlauthor{Hrayr Harutyunyan}{isi}
\icmlauthor{Greg Ver Steeg}{isi}
\icmlauthor{Aram Galstyan}{isi}
\end{icmlauthorlist}

\icmlaffiliation{isi}{Information Sciences Institute, University of Southern California}
\icmlaffiliation{goog}{Google AI, New York}

\icmlcorrespondingauthor{Sami}{sami@haija.org}
\icmlcorrespondingauthor{Bryan}{bperozzi@acm.org}

\icmlkeywords{Machine Learning, ICML}

\vskip 0.3in
]
\printAffiliationsAndNotice{}

\begin{abstract}
	Existing popular methods for semi-supervised learning with Graph Neural Networks (such as the Graph Convolutional Network)
	provably
	cannot learn a general class of neighborhood mixing relationships.
	To address this weakness, we propose a new model, \ourmodel, that can learn
	these relationships, including
	difference operators,
	by
	repeatedly mixing feature representations of neighbors at various distances.
	\ourmodel\ requires no additional memory or computational complexity, and outperforms on challenging baselines.
	In addition, we propose sparsity regularization that allows us to visualize how the network prioritizes neighborhood information across different graph datasets. Our analysis of the learned architectures reveals that neighborhood mixing varies per datasets. %

\end{abstract}

\section{Introduction}
Convolutional Neural Networks (CNNs) establish state-of-the-art performance for many Computer Vision applications \citep{alexnet, szegedy}. CNNs consist of a series of \textit{convolutional layers}, each parameterized by a \textit{filter} with pre-specified spatial dimensions.
CNNs are powerful because they are able to learn a hierarchy of translation invariant feature detectors. 

\begin{figure}[t!]
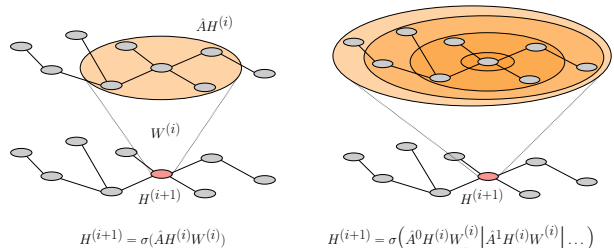

	\centering
	\begin{subfigure}[t]{0.25\textwidth}
		\centering
		\resizebox{100px}{!}{\input{figs/kipf.tex}}
		\caption{Traditional graph convolution. }
		\label{fig:kipf-feats}
	\end{subfigure}%
	\begin{subfigure}[t]{0.25\textwidth}
    	\centering
        \resizebox{100px}{!}{\input{figs/ourmodel.tex}}
		\caption{Our mixed feature model.}
		\label{fig:ourmodel-feats}
	\end{subfigure}
	\caption{
	Feature propagation in traditional graph convolution methods (a), compared to ours (b).
	We show the latent feature for red node in layer $i+1$ given node features in layer $i$. The traditional graph convolution case only aggregates from immediate neighbors $\hat{A}H^{(i)}$. In our \ourmodel\, the feature vector $H^{(i+1)}$ is a learned combination of the node's neighbors $\hat{A}^{j}H^{(i)}$ at multiple distances $j$.
	}
	\label{fig:crownjewel}
	\vspace{-0.4cm}
\end{figure}

The success of CNNs in Computer Vision and other domains has motivated researchers \citep{bruna,fast-spectrals,kipf} to extend the convolutional operator from regular grids, in which the structure is fixed and repeated everywhere, to graph-structured data, where nodes' neighborhoods can greatly vary in structure across the graph.
Generalizing convolution to graph structures should allow models to learn location-invariant node and neighborhood features. %

Early extensions of the graph convolution (GC) operator were theoretically motivated \citep{bruna}, but (1) required quadratic computational complexity in number of nodes and therefore were not scalable to large graphs, and (2) required the graph to be completely observed during training, targeting only the \textit{transductive} setting. 
\citet{fast-spectrals} and \citet{kipf} propose GC \textit{approximations} that are computationally-efficient (linear complexity, in the number of edges), and can be applied in \textit{inductive} settings, where the test graphs are not observed during training. 

However, said approximations limit the representational capacity of the model.
In particular, if we represent an image as a graph of pixel nodes, where edges connect adjacent pixels, GC approximations applied on the pixel graphs will be unable to learn Gabor-like\footnote{We use ``\textit{like}'', as graph edges are not axis-aligned.} filters.
Gabor filters are fundamental to the human visual cognitive system \citep{gabor, daugman85}. Further, these filters are automatically recovered by training CNNs on natural images (see \citet{alexnet, honglak} for visualizations).
Their automatic recovery implies their usefulness for hierarchical object representations and scene understanding, as guided by the optimization (e.g. classification) objective.
Since Graphs are generic data structures that can encode data from various domains (e.g. images, chemical compounds, social, and biological networks),
realizing Gabor-like filters in Graph domains ought to yield a general advantage.

In this work, we address the limitations of the approximations that prevent these models from capturing the graph analogue of Gabor filters.  Our proposed method, \ourmodel, allows full linear mixing of neighborhood information (as illustrated in Figure \ref{fig:crownjewel}), at every message passing step.
Specifically, our contributions are the following:
\begin{itemize}[itemsep=-3pt,topsep=-3pt]
    \item We formalize \textit{Delta Operators} and their generalization, \textit{Neighborhood Mixing}, to analyze the expressiveness of graph convolution models.  We show that popular graph convolution models (e.g.\ GCN of \citet{kipf}) cannot learn these representations.
    \item We propose \ourmodel, a new Graph Convolutional layer that mixes powers of the adjacency matrix. We prove that \ourmodel\ can learn a wider class of representations without increasing the memory footprint or computational complexity of previous GCN models.
    \item We provide a method of learning to divide modeling capacity among various widths and depths of a \ourmodel\ model, yielding powerful compact GCN architectures.  These architectures conveniently also allow visual inspection of which aspects of a graph are important.
\end{itemize}
We demonstrate our method on node classification tasks. Our code is available on \href{https://github.com/samihaija/mixhop}{github.com/samihaija/mixhop}.

\section{Preliminaries and Related Work}

\vspace{-0.2cm}
\subsection{Notation}
\vspace{-0.2cm}
Graph $G$ with $n$ nodes and $m$ edges has a feature matrix $X\in \mathbb{R}^{n \times {s_0}}$ with $s_0$ features per node, and training labels $Y_I$, annotating a partial set of nodes with the $c$ possible classes.  
The output of the task, $Y_O$, is an assignment of labels to the nodes, $Y \in [0,1]^{n \times c}$.
Let $A$ denote the adjacency matrix of $G$, where a non-zero entry $A_{ij}$ indicates an edge between nodes $i$ and $j$. 
We consider the case of a binary adjacency matrix ($A\in \{0,1\}^{n \times n}$), but this notation can be extended w.l.o.g.\ to weighted graphs.
Let $I_n$ be the $n\times n$ identity matrix, and $D$ be the degree matrix, $D = \textrm{diag}(d)$, where $d \in \mathbb{Z}^n$ is the degree vector with $d_j = \sum_i A_{ij}$.

\subsection{Message Passing}
\vspace{-0.2cm}
Message Passing algorithms can be used to learn models over graphs \cite{quantchem}. %
In such models, each graph node (and optionally edge) holds a latent vector, initialized to the node's input features
Each node repeatedly \textit{passes} its current latent vector to, and aggregates incoming messages from, its immediate neighbors. After $l$ steps of message passing and feature aggregation, every node outputs a representation which can be used for an upstream task e.g. node classification, or entire graph classification. The $l$ steps (message passing and aggregation) can be parametrized
and trained via Backprop-Through-Structure algorithms \cite{bpts}, to minimize an objective measured using the node representations as output by the $l$'th step.

\subsection{Graph Convolutional Networks}
We refer to the Graph Convolutional Network proposed by \citet{kipf} as the \textit{vanilla GCN}. The vanilla GCN Graph Convolutional (GC) Layer is defined as:
\begin{equation}
\label{eq:gclayer}
H^{(i+1)} = \sigma(\widehat{A} H^{(i)} W^{(i)}),
\end{equation}
where  $H^{(i)} \in \mathbb{R}^{n \times s_{i}}$ and $H^{(i+1)} \in \mathbb{R}^{n \times s_{i+1}}$  are the input and output activations for layer $i$,
$W^{(i)} \in \mathbb{R}^{s_i \times s_{i+1}}$ is a trainable weight matrix and $\sigma$ is an element-wise activation function, and $\widehat{A}$ is a symmetrically normalized adjacency matrix with self-connections, 
$\widehat{A} = D^{-\frac12} (A+I_n) D^{-\frac12}$. 
A GCN model with $l$ layers is then defined as:
\begin{equation}
  H^{(i)} =
    \begin{cases}
      X & \text{if $i = 0$} \\
      \sigma(\widehat{A} H^{(i-1)} W^{(i-1)}) & \text{if $i \in [1$ .. $l]$},
    \end{cases}
    \label{eq:kipfmodel}
\end{equation}
and the output $Y_O$ can be set as to function of $H^{(l)}$.
The vanilla GCN can be described as a message passing algorithm, where a node's latent representation at step $i$ is defined as an \textit{average} of its neighbors' representations from step $i-1$, multiplied by $W^{(i-1)}$. See \citet{quantchem}.
The vanilla GCN makes three simplifying assumptions: (1) it is a Chebyshev rank-2 approximation of multiplication in the Graph Fourier basis, defined to be the eigenbasis of the graph Laplacian; (2) it assumes that the two coefficients of the Chebyshev polynomials multiply to -1; (3) a \textit{renormalization trick} adds self-connections (identity matrix) to $A$ before, rather than after, normalization.
These simplifications reduce the computational complexity and prevent exploding/vanishing gradients.
However, it simplifies the definition of convolution to become a simple neighborhood-averaging operator: this is obvious from Equation \ref{eq:gclayer} -- the features are left-multiplied by normalized adjacency $\widehat{A}$, effectively replacing each row in the feature matrix, by the average of its neighbors (and itself, due to \textit{renormalization}).

\subsection{Semi-supervised Node Classification}
We are interested in semi-supervised node classification tasks.
To train a GCN model on such a task, we select row slices from the output matrix $Y_O$, corresponding to nodes with known labels in $Y_I$, on which a loss and its gradients are evaluated. The gradient of the loss is backpropagated
through the GC layers where they get multiplied by $\widehat{A}^\top$, spreading gradients to unlabeled examples.

\section{Our Proposed Architecture}
\label{sec:architecture}

We are interested in higher-order message passing, where nodes receive latent representations from their immediate (first-degree) neighbors and from further N-degree neighbors at every message passing step. 
In this section, we motivate and detail a model with trainable aggregation parameters that can choose how to mix latent information from neighbors at various distances. 
Our analysis starts with the \emph{Delta Operator}, a subtraction operation between node features collected from different distances. The vanilla GCN is unable to learn such a feature representation.
Before introducing our model, we give one formal definition:

\begin{definition}
	\label{def:twohopdelta}
	\underline{Representing Two-hop Delta Operator}: A model is capable of representing a two-hop Delta Operator if there exists a setting of its parameters and an injective mapping $f$, such that the output of the network becomes%
	\begin{equation}
	\label{eq:deltaop}
	f \left( \sigma \left(\widehat{A} X \right) - \sigma \left( \widehat{A}^2 X \right)\right),
	\end{equation}
    given \textbf{any} adjacency matrix $\widehat{A}$, features $X$, and activation function $\sigma$.
\end{definition}

Learning such an operator should allow models to represent feature differences among neighbors, which is necessary, for example, for learning Gabor-like filters on the graph manifold.
To provide a concrete example regarding graphs, consider an online social network.
In this setting, Delta Operators  allow a model to represent users that live around the ``boundary'' of social circles \citep{Perozzi:2018:DCA:3178544.3139241}.
To learn an approximate feature for \textit{American person with a popular German friend}, who might have most immediate friends speaking English, but many friends-of-friends speaking German. This person can be represented by learning a convolutional filter contrasting the English and German languages of one-hop and two-hop neighbors.

Note that in the Definition \ref{def:twohopdelta} we allow not learning the direct form of two-hop Delta Operators, but a transformation of it, as long as that transformation can be inverted (i.e. $f$ is injective).

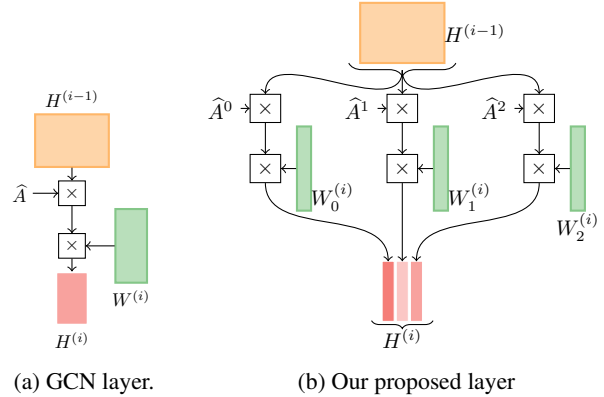
\begin{figure}[t!]
	\centering
	\begin{subfigure}[b]{0.15\textwidth}
		\centering
		\begin{tikzpicture}[scale=0.7, transform shape]
\definecolor{paramcolor}{HTML}{50ae55}
\definecolor{inputcolor}{HTML}{fd9727}
\definecolor{outputcolor}{HTML}{f1453d}
\tikzset{
	inputmat/.style = {rectangle, draw=inputcolor!70, fill=inputcolor!40, thick, minimum width=1.4cm, minimum height = 1cm},
	outputmat/.style = {rectangle, draw=outputcolor!0, fill=outputcolor!50, thick, minimum width=0.6cm, minimum height = 1.0cm},
	thickermat/.style = {rectangle, draw=black!50, fill=black!20, thick, minimum width=1.4cm, minimum height = 1cm},
	parameter/.style = {rectangle, draw=paramcolor!70, fill=paramcolor!40, thick, minimum width=0.6cm, minimum height = 1.4cm},
}
\node [inputmat](x) [draw] { }; 
\node [above =0cm of x] (xlabel) {$H^{(i-1)}$ }; 
%\node [below of=x](xlabel) {$H^{(l)}$}
\node [draw, below of=x](mult) {$\times$};
\node [left of=mult] (a) {$\widehat{A}$};
%\node [thickermat, below of=mult](ax) [draw] {}; 
\node [draw, below of=mult](wmult) {$\times$};
\node [parameter, right =0.55cm of wmult](w) {};
\node [below=0cm of w](wl) {$W^{(i)}$}; % 
\node [outputmat, below of=wmult](y) {};
\node [below =0cm of y] (xlabel) {$H^{(i)}$ }; 
%\node [draw, right of=wmult](sigma) {$\sigma$};

\draw[-{>}] (x) -- (mult);
\draw[-{>}] (a) -- (mult);
%\draw[-{>}] (mult) -- (ax);
\draw[-{>}] (mult) -- (wmult);
\draw[-{>}] (w) -- (wmult);
\draw[-{>}] (wmult) -- (y);
\end{tikzpicture}
		\caption{GCN layer.}
	\end{subfigure}%
	\begin{subfigure}[b]{0.35\textwidth}
		\centering
		\begin{tikzpicture}[scale=0.8, transform shape]
\definecolor{paramcolor}{HTML}{50ae55}
\definecolor{inputcolor}{HTML}{fd9727}
\definecolor{outputcolor}{HTML}{f1453d}

\tikzset{
	inputmat/.style = {rectangle, draw=inputcolor!70, fill=inputcolor!40, thick, minimum width=1.4cm, minimum height = 1cm},
	output1mat/.style = {rectangle, draw=black!0, fill=outputcolor!70, line width=0, inner sep=0pt, minimum width=0.2cm, minimum height = 1.0cm},
	output2mat/.style = {rectangle, draw=black!0, fill=outputcolor!30, line width=0, inner sep=0pt, minimum width=0.2cm, minimum height = 1.0cm},
	output3mat/.style = {rectangle, draw=black!0, fill=outputcolor!50, line width=0, inner sep=0pt, minimum width=0.2cm, minimum height = 1.0cm},
	thickermat/.style = {rectangle, draw=black!50, fill=black!20, thick, minimum width=1.4cm, minimum height = 1cm},
	thinmat/.style = {rectangle, draw=black!50, fill=black!20, thick, minimum width=0.2cm,
		minimum height = 1cm},
	parameter/.style = {rectangle, draw=paramcolor!70, fill=paramcolor!40, thick, minimum width=0.2cm, minimum height = 1.4cm},
}
\node [inputmat,anchor=south](x) [draw] {}; 
\node [right=-0.4em of x] (xlabel) {$H^{(i-1)}$};

%\node [below of=x](xlabel) {$H^{(l)}$}

\draw [decorate,decoration={brace,amplitude=5pt,mirror,raise=1ex}]
  ($(x.south west)+(-0.5em, 0.6em)$) -- ($ (x.south east)+(0.5em, 0.6em) $) node[midway,yshift=0em]{};

\node [draw, below left=0.5cm and 1.3cm of x](mult) {$\times$};
\node [left=0.15cm of mult] (a) {$\widehat{A}^0$};
%\node [thickermat, below of=mult](ax) [draw] {}; 
\node [draw, below of=mult](wmult) {$\times$};
\node [parameter, right =0.25cm of wmult](w1) {};
\node [below right=-0.6cm and -0.14cm of w1](wl) {$W^{(i)}_0$};
%\node [output1mat, below of=wmult](y1) {};

\node [draw, below=0.5cm of x](mult2) {$\times$};
\node [left=0.15cm of mult2] (a2) {$\widehat{A}^1$};
%\node [thickermat, below of=mult2](a2x) [draw] {}; 
\node [draw, below of=mult2](wmult2) {$\times$};
\node [parameter, right =0.25cm of wmult2](w2) {};
\node [below right=-0.6cm and -0.14cm of w2](wl) {$W^{(i)}_1$};
%\node [output2mat, below of=wmult2](y2) {};

\node [draw, below right=0.5cm and 1.3cm of x](mult3) {$\times$};
\node [left=0.15cm of mult3] (a3) {$\widehat{A}^2$};
%\node [thickermat, below of=mult3](a3x) [draw] {}; 
\node [draw, below of=mult3](wmult3) {$\times$};
\node [parameter, right =0.25cm of wmult3](w3) {};
\node [below=-0.1cm of w3](wl) {$W^{(i)}_2$};
%\node [output3mat, below of=wmult3](y3) {};

%\node [draw, below of=y2](concat) {concatenate};

\node [output2mat, below=1.3cm of wmult2](yy2) {};
\node [output1mat, left=0.1em of yy2](yy1) {};
\node [output3mat, right=0.1em of yy2](yy3) {};
\node [below=-0em of yy2] (xlabel) {$H^{(i)}$};

\draw [decorate,decoration={brace,amplitude=3pt,mirror,raise=1ex}]
  ($(yy1.south west)+(-0.5em, 0.6em)$) -- ($ (yy3.south east)+(0.5em, 0.6em) $) node[midway,yshift=0em]{};

\draw ($(x.south) + (0em, -0.3em)$) edge[out=270,in=90,->] (mult);
\draw[-{>}] (a) -- (mult);
\draw[-{>}] (mult) -- (wmult);
\draw[-{>}] (w1) -- (wmult);
\draw (wmult) edge[out=270,in=90,->] (yy1);

\draw ($(x.south) + (0em, -0.3em)$) edge[out=270,in=90,->] (mult2);
\draw[-{>}] (a2) -- (mult2);
%\draw[-{>}] (mult2) -- (a2x);
\draw[-{>}] (mult2) -- (wmult2);
\draw[-{>}] (w2) -- (wmult2);
\draw  (wmult2) edge[out=270,in=90,->] (yy2);

\draw ($(x.south) + (0em, -0.3em)$) edge[out=270,in=90,->] (mult3);
\draw[-{>}] (a3) -- (mult3);
%\draw[-{>}] (mult3) -- (a3x);
\draw[-{>}] (mult3) -- (wmult3);
\draw[-{>}] (w3) -- (wmult3);
\draw  (wmult3) edge[out=270,in=90,->] (yy3);
%\draw[-{>}] (y1) -- (yy1);
%\draw[-{>}] (y2) -- (yy2);
%\draw[-{>}] (y3) -- (yy3);
%\draw[-{>}] (concat) -- (yy2);
\end{tikzpicture}
		\caption{Our proposed layer}
		\label{fig:ourmodel}
	\end{subfigure}		\caption{
		Vanilla GC layer (a), using adjacency $\widehat{A}$, versus our GC layer (b), using powers of $\widehat{A}$. Orange denotes an input activation matrix, with one row per node; green denotes the trainable parameters; and red denotes the layer output.
		Left vs right-multiplication is specified by the relative position of the multiplicand to the $\times$ operator.
	}
	\label{fig:gc_layer_comparison}
\end{figure}

In Sections \ref{sec:mixhop} - \ref{sec:representcap}, we analyze the extent to which various GCN models can learn the Delta Operator. We generalize this definition and analysis in Section \ref{sec:generalizing}.

\subsection{MixHop Graph Convolution Layer}
\label{sec:mixhop}
We propose replacing the Graph Convolution (GC) layer defined in Equation \ref{eq:gclayer}, with:
\begin{equation}
\label{eq:ourlayer}
H^{(i+1)} =\underset{j \in P}{\Bigg\Vert} \sigma\left(\widehat{A}^j H^{(i)} W_j^{(i)}\right),
\end{equation}
where the hyper-parameter $P$ is a set of integer adjacency powers, $\widehat{A}^j$ denotes the adjacency matrix $\widehat{A}$ multiplied by itself $j$ times, and $\Vert$ denotes column-wise concatenation. 
The difference between our proposed layer and a vanilla GCN is shown in Figure \ref{fig:gc_layer_comparison}.
Note that setting $P=\{1\}$ exactly recovers the original GC layer. Further, note that $\widehat{A}^0$ is the identity matrix $I_{n}$, where $n$ is the number of nodes in the graph. We depict a model with $P=\{0, 1, 2\}$ in Figure \ref{fig:ourmodel-feats}. %
In our model, each layer contains $|P|$ distinct parameter matrices, each of which can be a different size. By default, we set all $|P|$ matrices to have the same dimensionality;
however, in Section \ref{sec:meta}, we explain how we utilize sparsifying regularizers on the learnable weight matrices to produce dataset-specific model architectures that slightly outperform our default settings.

\subsection{Computational Complexity}
There is no need to calculate $\widehat{A}^j$. We calculate  $\widehat{A}^j H^{(i)}$ with right-to-left multiplication. Specifically, if $j=3$, we calculate $\widehat{A}^3 H^{(i)}$ as $ \widehat{A} \left(\widehat{A} \left(\widehat{A} H^{(i)} \right) \right) $. Since we store $\widehat{A}$ as a sparse matrix with $m$ non-zero entries, 
an efficient implementation of our layer (Equation \ref{eq:ourlayer}) takes $\mathcal{O}(j_\textrm{max} \times m \times s_i)$ computational time, where $j_\textrm{max}$ is the largest element in $P$ and $s_i$ is the feature dimension of $H^{(i)}$. Under the realistic assumptions of $j_\textrm{max} \ll m$ and $s_l \ll m$, running an $l$-layer model takes $\mathcal{O}(lm)$ computational time. This matches the computational complexity of the vanilla GCN.

\begin{algorithm}[tb]
   \caption{MixHop Graph Convolution Layer}
   \label{alg:example}
\begin{algorithmic}
   \STATE {\bfseries Inputs:} $H^{(i-1)}$, $\widehat{A}$ 
   \STATE {\bfseries Parameters:} $\{W_j^{(i)}\}_{j \in P}$
   \STATE $j_{\text{max}} := \max P$
   \STATE $B := H^{(i-1)}$
   \FOR{$j=1$ {\bfseries to} $j_{\text{max}}$}
        \STATE $B:= \widehat{A} B$
        \IF{$j \in P$}
            \STATE $O_j := B W_j^{(i)}$
        \ENDIF
   \ENDFOR
   \STATE $H^{(i)} := \underset{j \in P}{\Vert}O_j$
   \STATE {\bfseries Return:} $H^{(i)}$
\end{algorithmic}
\end{algorithm}

\subsection{Representational Capability}
\label{sec:representcap}
Since each layer outputs the multiplication of different adjacency powers in different columns, the next layer's weights can learn arbitrary linear combinations of the columns. By assigning a positive coefficient to a column produced by some $\widehat{A}$ power, and assigning a negative coefficient to another, the model can learn a Delta Operator. In contrast, vanilla GCNs are not capable of representing this class of operations, even when stacked over multiple layers.

\begin{theorem}
    \label{theorem:kipf}
    The vanilla GCN defined by Equation \ref{eq:kipfmodel} is \textbf{not} capable of representing two-hop Delta Operators.
\end{theorem}

\begin{theorem}
	\label{theorem:ours}
    MixHop GCN (using layers defined in Equation \ref{eq:ourlayer}) can represent two-hop Delta Operators.
\end{theorem}

\textbf{Proof of Theorem \ref{theorem:kipf}.}
The output of an $l$-layer vanilla GCN has the following form:
\begin{equation*}
    \sigma(\widehat{A} (\sigma(\widehat{A}\cdots \sigma(\widehat{A} X W^{(0)}) \cdots )W^{(l-2)}) W^{(l-1)}).
\end{equation*}
For the simplicity of the proof, let's assume that $\forall i, s_i = n$.
In a particular case, when $\sigma(x) = x$ and $X = I_n$, this reduces to $\widehat{A}^l W^*$, where $W^* = W^{(0)}W^{(1)}\cdots W^{(l-1)}$.
Suppose the network is capable of representing a two-hop Delta Operator. This means that there exists an injective map $f$ and a value for $W^*$, such that $\forall \widehat{A}, \widehat{A}^l W^* = f(\widehat{A}  - \widehat{A}^2 )$. Setting $\widehat{A} = I_n$, we get that $W^* = f(0)$.
Let
\begin{equation*}
\widehat{C}_{1,2} \triangleq \begin{bmatrix}[ccccc]0.5 & 0.5 & 0 & \cdots & 0\\
0.5 & 0.5 & 0 & \cdots & 0\\
0 & 0 & 1 & \cdots & 0\\
\vdots & \vdots & \vdots & \ddots & \vdots \\
0 & 0 & 0 & \cdots & 1
\end{bmatrix}
\end{equation*}
be the symmetrically normalized adjacency matrix with self-connections corresponding to the graph having a single edge between vertices 1 and 2. Setting $\widehat{A} = \widehat{C}_{1,2}$, we get $\widehat{C}_{1,2}W^* = f(0)$. Since we already have that $f(0) = W^*$, we get that $(I_n - \widehat{C}_{1,2})W^* = 0$, which proves that the $w^*_1 = w^*_2$, where $w^*_i$ is the $i$-th row of $W^*$. Since the choice of vertices 1 and 2 was arbitrary, we have that all rows of $W^*$ are equal to each other. Therefore, $\text{rank}(\widehat{A}^l W^*)\le 1$, which implies that outputs of mapping $f$ should be at most rank-one matrices. Thus, $f$ cannot be injective, proving that vanilla GCN cannot represent two-hop Delta Operators.
\\
\null\hfill $\blacksquare$

\textbf{Proof of Theorem \ref{theorem:ours}.}
A two-layer model, defined using Equation \ref{eq:ourlayer} with $P=\{0, 1, 2\}$ recovers the two-hop delta operator defined in Equation \ref{eq:deltaop}. We start by redefining the feature vector $H^{(1)}$ learned by the first layer of the model by pulling out the element-wise activation function $\sigma$ and expanding the concatenation operator found in the layer definition:
\begin{align*}
H^{(1)} &= \underset{j \in \{0, 1, 2\}}{\Bigg\Vert} \sigma\left(\widehat{A}^j X W_j^{(0)}\right) \\
		&= \sigma\left( \underset{j \in \{0, 1, 2\}}{\Bigg\Vert} \widehat{A}^j X W_j^{(0)}  \right) \\
            &=  \sigma\left( \left[ \begin{array}{c;{2pt/2pt}c;{2pt/2pt}c}
				I_N X W_0^{(0)} &
				\widehat{A} X W_1^{(0)} & \widehat{A}^2 X W_2^{(0)}
			\end{array} \right] \right),
\end{align*}
We can now set $W_0^{(0)}=0$ (zero matrix) and $W_1^{(0)}=W_2^{(0)} = I_{s_0}$. The expression above can be simplified to $H^{(1)} = \sigma\left(\left[\begin{array}{c;{2pt/2pt}c;{2pt/2pt}c}
0 &
\widehat{A} X  & \widehat{A}^2 X
\end{array}\right] \right)$. The feature vector $H^{(1)}$  can be plugged into the equation for the second layer that has linear activation function:
\begin{align*}
H^{(2)} &= \left[\begin{array}{c;{2pt/2pt}c;{2pt/2pt}c}
I_N H^{(1)} W_0^{(1)} &
\widehat{A} H^{(1)} W_1^{(1)} & \widehat{A}^2 H^{(1)} W_2^{(1)}
\end{array}\right].
\end{align*}
Setting the weights for the second layer as $W_1^{(1)} = W_2^{(1)} = 0$, and
\begin{equation}
W_0^{(1)}=\begin{bmatrix}[c]  0 \\ I_{s_0} \\ -I_{s_0}  \end{bmatrix},
\end{equation}
makes 
\begin{align*}
H^{(2)} = \left[\begin{array}{c;{2pt/2pt}c;{2pt/2pt}c}\left(\sigma\left(\widehat{A} X\right) - \sigma \left(\widehat{A}^2 X \right) \right) &
0 & 0
\end{array}\right].
\end{align*}

This shows that our GCN can successfully represent the two-hop Delta Operators according to the Definition \ref{def:twohopdelta}.
\null\hfill $\blacksquare$

\subsection{General Neighborhood Mixing}
\label{sec:generalizing}
We generalize Definition \ref{def:twohopdelta} from two-hops to multiple hops:
\begin{definition}
	\label{def:general}
	\underline{General layer-wise Neighbor-hood Mixing}:
	A Graph Convolutional Network is capable of representing layer-wise neighborhood mixing if for any $\alpha_0, \alpha_1, \ldots, \alpha_m$ numbers, there exists a setting of its parameters and an injective mapping $f$, such that the output of the network becomes equal to
	\begin{equation}
	\label{eq:general}
	f\left(\sum_{j=0}^m \alpha_j \sigma\left( \widehat{A}^j X \right)\right) 
	\end{equation}
	for \textbf{any} adjacency matrix $\widehat{A}$, features $X$, and activation function $\sigma$.
\end{definition}

\begin{theorem}
	\label{theorem:generalizedkipf}
	GCNs defined using Equation \ref{eq:gclayer} are \textbf{not} capable of representing general layer-wise neighborhood mixing.
\end{theorem}
\begin{theorem}
	\label{theorem:generalizedours}
	GCNs defined using our proposed method (Equation \ref{eq:ourlayer}) are capable of representing general layer-wise neighborhood mixing.
\end{theorem}

\textbf{Proof of Theorem \ref{theorem:generalizedkipf}}.
This trivially follows from Theorem \ref{theorem:kipf}: if the vanilla GCN cannot recover a two-hop Delta Operator, defined in Equation \ref{eq:deltaop}, it cannot recover the Delta Operator generalization in Equation \ref{eq:general}. 
\null\hfill $\blacksquare$

\textbf{Proof of Theorem \ref{theorem:generalizedours}.}
The proof steps closely resemble the proof of Theorem \ref{theorem:ours}. Our GCN with $P=\{0, \dots, m\}$ can represent the target function, by setting the first layer weight matrices as $W^{(0)}_j =  I_{s_0}, \ \forall j \in P$ and setting all but the zeroth second layer weight matrices as $W^{(1)}_1 = W^{(1)}_2 = \dots = W^{(1)}_m = 0$. In other words, we utilize only zero-hops in the second layer, setting the zeroth-power weight matrix the following way:
\begin{equation}
W^{(1)}_0 = \left[ \begin{array}{c} \alpha_0 I_{s_0} \\ \vdots \\ \alpha_m I_{s_0} \end{array} \right]
\end{equation}
This setting of parameters exactly recover the expression in Equation \ref{eq:general}, for any adjacency matrix $\widehat{A}$ and features $X$.
\null\hfill $\blacksquare$

We note that the generalized Delta Operator in Definition \ref{def:general} does not explicitly specify feature differences as in Definition \ref{def:twohopdelta}; rather, the generalized form defines linear combinations of features (which includes subtraction). 

\section{Learning Graph Convolution Architectures}

We have discussed a single layer of our model. In practice, one would stack multiple layers and interleave them with standard neural operators such as BatchNorm \cite{batchnorm}, element-wise activation, and Dropout \cite{dropout}. In this section, we discuss approaches to turning the \ourmodel\ GC layer into a \ourmodel\ GCN.

\subsection{Output Layer}
\label{sec:outputlayer}
The final layer of a GCN performs a key role
for learning
the learned latent space of the model
on the dataset that is being trained on. %
As \ourmodel\ uniquely mixes features from different sets of information, we theorized that constraining the output layer may result in better outcomes for different tasks.
In order to leverage this property, we define our output layer in the following way:
We divide $s_l$ columns into sets of size $c$ and compute $\widetilde{Y}_O = \sum_{k=1}^{s_l/c} q_k H^{(l)}_{*,(i d_l/c\ :\ (i+1)s_l/c)}$, then $Y_O = \textrm{softmax}(\widetilde{Y}_O)$. Here the subscript on $H^{(l)}$ selects $c$ contiguous columns and the scalars $q_k \in [0, 1]$ define a valid distribution (output of a softmax). 
This results in the model being forced to choose which features it wants to prioritize by putting more weight on that feature. We obtain the model parameters $W_i^{(j)}$ for all $i,j$ and $q_1, \dots q_{\frac{s_l}{c}}$, by minimizing cross-entropy loss, measured only on nodes with known labels i.e. similar to \citep{kipf}.

\subsection{Learning Adjacency Power Architectures}
\label{sec:meta}
As mentioned, our model learns multiple weight matrices $W_j^{(i)}$, one per adjacency power used in the model. By default, we set all $W_j^{(i)}$ to be the same size, which effectively assigns the same capacity to adjacency powers $\widehat{A}^j$ for all $j \in P$. We intuit that different sizes of $W_j^{(i)}$ may be more appropriate for different tasks and datasets; as such, we are interested in learning how to automatically size $W_j^{(i)}$.

For vanilla GCNs, such an architecture search is relatively inexpensive - the parameters are the number of layers and their widths.
In contrast, searching over the architecture space of our model is multiplicatively  $\mathcal{O}(l \times |P|)$ more expensive, as each architecture involves choices on how to divide each layer width $s_i$ among the adjacency powers. %
To address this limitation, we propose using a lasso regularization to automatically learn an architecture for our model \citep{morphnet}. In particular, we train our architecture in stages:
\begin{enumerate}[nosep]
    \item Construct a wide network (e.g. 200 dimensions for each adjacency power, at each layer), only making choices on the depth. %
    \item Train the network on the task while applying L2 Group Lasso regularization over each column of each $W_j^{(l)}$. This will drop values of entire columns (close) to zero.
    \item At the peak validation accuracy, measure the L2 norm of each $W_j^{(l)}$. Pick a threshold, and count the number of columns in each $W_j^{(l)}$ with norm higher than the threshold. In our experiments, we pick a threshold such that the size of the shrunken model equals size of our baseline model (i.e. with $P=\{1\}$).
    \item Shrink the weight matrices by removing columns with norms below the $k$'th percentile. 
    \item Substitute L2 Group Lasso with standard L2 regularization. Restart training.

\end{enumerate}
We discuss the learned architectures in Section \ref{sec:expmeta}.

\section{Experimental Design}
\setlength{\tabcolsep}{0.3em}
\begin{table*}
	\begin{center}
		\begin{tabular}[t]{l|c|c|c}
			\textbf{Model}  & \textbf{Citeseer} & \textbf{Cora} & \textbf{Pubmed}  \\
			\hline
			ManiReg \citep{manireg}& $60.1$ & $59.5$ & $70.7$ \\ 
			
			SemiEmb \citep{semiemb}& $59.6$ & $59.0$ & $71.1$ \\ 
			
			LP \citep{lp}& $45.3$ & $68.0$ & $63.0$ \\ 
			
			DeepWalk \citep{deepwalk}& $43.2$ & $67.2$ & $65.3$ \\ 
			
			ICA \citep{ica}& $69.1$ & $75.1$ & $73.9$ \\ 
			
			Planetoid \citep{planetoid}& $64.7$ & $75.7$ & $77.2$ \\ 
			
			Vanilla GCN \citep{kipf}& $70.3$ & $81.5$ & $79.0$ \\ 
			\hline
			\hline

MixHop with $P=\{1\}$ (baseline) & $70.7 \pm$$0.73 $ & $ 81.1 \pm$$0.84 $ & $79.9 \pm$$0.78 $\\ 

MixHop: default architecture (ours) & $\mathbf{71.4} \pm$$0.81 $ & $81.8 \pm$$0.62 $ & $ 80.0 \pm$$1.1 $\\ 

MixHop: learned architecture (ours) & $\mathbf{71.4} \pm$$0.81 ^{\dagger}$ & $\mathbf{81.9} \pm$$0.40 $ & $\mathbf{80.8} \pm$$0.58 $\\ 
\hline
		\end{tabular}
	\end{center}
    \caption{Experiments run on Node Classification citation datasets created by \citet{planetoid}.  $^{\dagger}$The learned architecture for Citeseer is equivalent to default architecture, so the results are the same.}
\end{table*}

\begin{table}[t]
    \centering
	    \begin{tabular}[t]{l|c|c|c|c|c|c}
		\textbf{Dataset} & nodes & edges & features & $c$ & $|Y_I^P|$ &$|Y_I^R|$  \\
        \hline
        Citeseer & 3,327 & 4,732 & 3,703 & 6 & 120  & 600 \\
        Cora & 2,708 &5,429 & 1,433 & 7 & 140 & 700 \\
        Pubmed & 19,717 &44,338 & 500  & 3 & 60 & 300 \\
        \end{tabular}
        \caption{Dataset statistics. Numbers of nodes $(n)$, edges $(m)$, features, classes $(c)$, and labeled nodes ($|Y_I^P|$ from the Planetoid splits,  $|Y_I^R|$ from our random splits).}
        \label{tab:planetoid_stats}
\end{table}

\begin{table*}
	\begin{center}
		\begin{tabular}[t]{l|c|c|c}
			\textbf{Model}  & \textbf{Citeseer} & \textbf{Cora} & \textbf{Pubmed}  \\
			2-Layer MLP & $70.6 \pm$$1$ & $69.0 \pm$$1.1$ & $78.3 \pm$$0.54$ \\
			Chebyshev \citep{fast-spectrals} & $74.2 \pm$$0.5$ & $85.5 \pm$$0.4$ & $81.8 \pm$$0.5$ \\
            Vanilla GCN \citep{kipf}& $76.7 \pm$$0.43$ & $86.1 \pm$$0.34$ & $82.2 \pm$$0.29$ \\
            GAT \citep{gat} & $74.8 \pm$$0.42$ & $83.0 \pm$$1.1$ &  $81.8 \pm$$0.18$ \\  
			\hline
			\hline
			MixHop: default architecture (ours)
			& $76.3 \pm$$0.41$ & $87.0 \pm$$0.51$ & $83.6 \pm$$0.68$ \\
			MixHop: learned architecture (ours)
			& $\mathbf{77.0} \pm$$0.54$ & $\mathbf{87.2} \pm$$0.32$ & $\mathbf{83.8} \pm$$0.44$ \\
		\end{tabular}
    \end{center}
    \caption{Classification results on random partitions of \citep{planetoid} datasets.}
    \label{tab:realistic}
\end{table*}

Given the model described above, a number of natural questions arise.
In this section, we aim to design experiments which answer the following hypothesises:
\begin{itemize}[topsep=-4pt,itemsep=-3pt]
    \item \textbf{H1}: The \ourmodel\ model learns delta operators.
    \item \textbf{H2}: Higher order graph convolutions using neighborhood mixing can outperform existing approaches (e.g. vanilla GCNs) on real semi-supervised learning tasks. 
    \item \textbf{H3}: When learning a model architecture for \ourmodel\, the best performing architectures differ for each graph.
\end{itemize}

To answer these questions, we design three experiments.
\begin{itemize}[topsep=-4pt,itemsep=-3pt]
    \item \textbf{Synthetic Experiments}: This experiment uses a family of synthetic graphs which allow us to vary the correlation (or \emph{homophily}) of the edges in a generated graph, and observe how different graph convolutional approaches respond.  
    As homophily is decreased in the network, nodes are more likely to connect to those with different labels, and a model that better captures delta operators should have superior performance.
    \item \textbf{Real-World Experiments}:  This experiment evaluates \ourmodel's performance on a variety of noisy real world datasets, comparing against challenging baselines.
    \item \textbf{Model Visualization Experiment}: 
    This experiment shows how an appropriately regularized \ourmodel model can learn different, task-dependent, architectures.
\end{itemize}
\vspace{-0.1cm}
\subsection{Datasets}
We conduct semi-supervised node classification experiments on synthetic and real-world datasets.

\textbf{Synthetic Datasets}:
Our synthetic datasets are generated following \citet{karimi}. We generate 10 graphs, each with a different homophily coefficient (ranging from 0.0 to 0.9 at 0.1 intervals) that indicates the likelihood of a node forming a connection to a neighbor with the same label.
For example, a node in the $homophily=0.9$ graph with 10 edges, will have on average 9 edges to a same-label neighbor. All graphs contain 5000 nodes.
The features for all synthetic nodes were sampled from overlapping multi-Gaussian distributions. We randomly partition each graph into train, test, and validation node splits, all of equal size. See Appendix for more information.

\textbf{Real World Datasets}:
The experiments with real-world datasets follow the methodology proposed in \citet{planetoid}.  In addition to using the classic dataset split, (which have 20 samples per label), we evaluate against against a set of random splits with 100 samples per label. We will release our test splits.

\subsection{Training}
For all experiments, we construct a 2-layer network of our model using TensorFlow \citep{tensorflow}. We train our models using a Gradient Descent optimizer for a maximum of 2000 steps, with an initial learning rate of 0.05 that decays by 0.0005 every 40 steps. We terminate training if validation accuracy does not improve for 40 consecutive steps; as a result, most runs finish in less than 200 steps. We use $5\times10^{-4}$ L2 regularization on the weights, and dropout input and hidden layers. We note that the citation datasets are extremely sensitve to initializations; %
as such, we run all models 100 times, sort by the validation accuracy, and finally report the test accuracy for the top 50 runs. For all models we ran (our models in Tables 1 \& 3, and all models in Table 3), we use a latent dimension of 60;
Our default architecture evenly divided 60 dimensions are divided evenly to all $|P|$ powers. Our learned architectures spread them unevenly, see Section \ref{sec:expmeta}.

\section{Experimental Results}

\subsection{Results on Synthetic Graphs}

\begin{figure}[t!]
	\centering
	\includegraphics[width=0.9\linewidth,trim={0 0 0 0cm},clip=true]{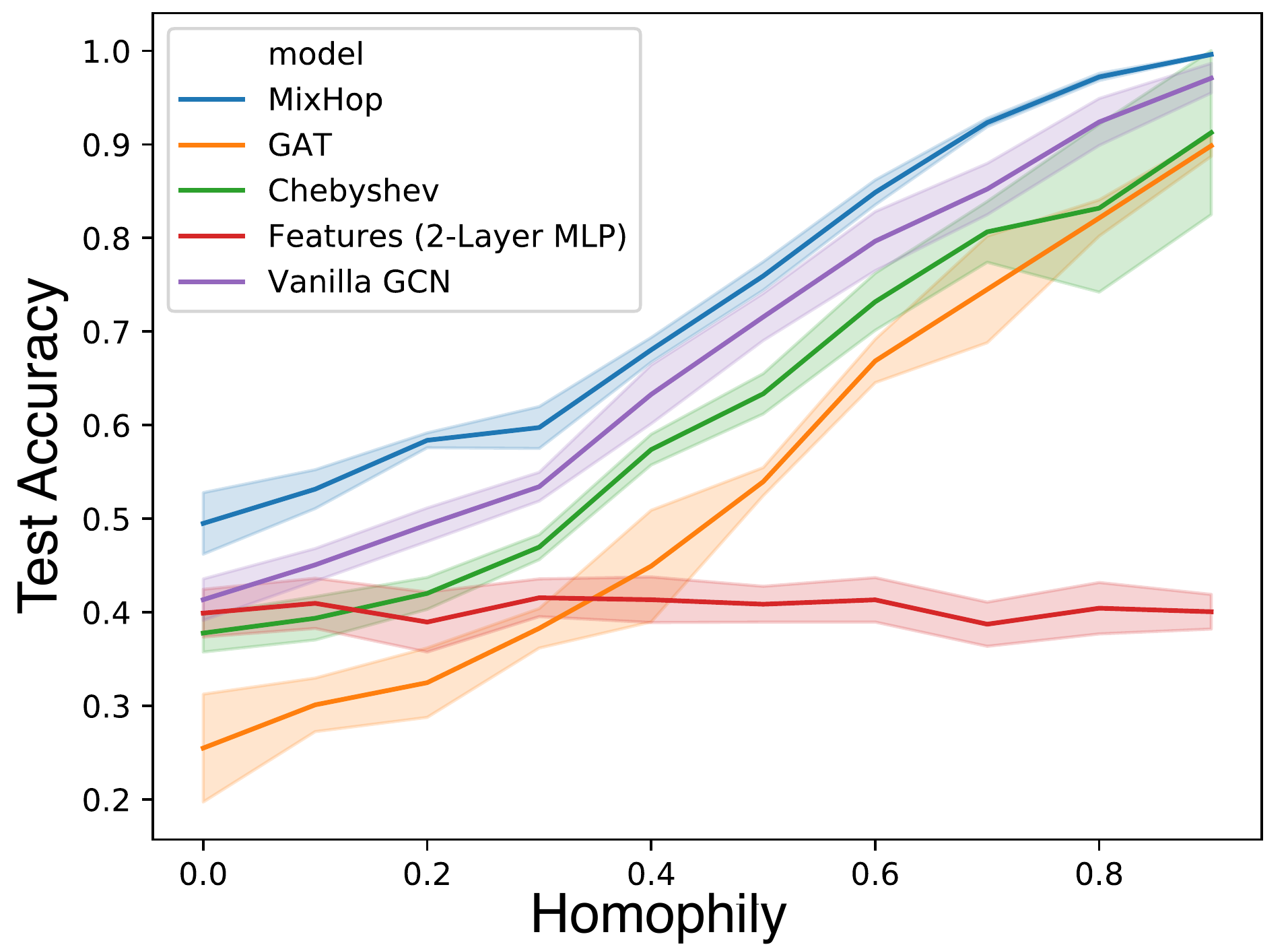}
	\caption{Synthetic dataset results.
		MLP does not utilize graph with (homophilic) edges, but only node features.}
	\label{fig:synthetic}
	\vspace{-0.2cm}
\end{figure}

We present our results on the synthetic datasets in Figure \ref{fig:synthetic}. We show average accuracy for each baseline against the homophily of the graph. We use a dense (MLP) model that does not ingest any adjacency information as a control. As expected, all models perform better as the homophily of the synthetic graph increases. At low levels of homophily, when nodes are rarely adjacent to neighbors with the same label, we observe that \ourmodel\ performs significantly better than the most competitive baseline. Interestingly, we notice that the GAT model performs significantly worse than the features-only control. This suggests that the added attention mechanism of the GAT model relies heavily on homophily in node neighborhoods.

\begin{figure}
	\centering
	\centering
	\includegraphics[width=0.8\linewidth]{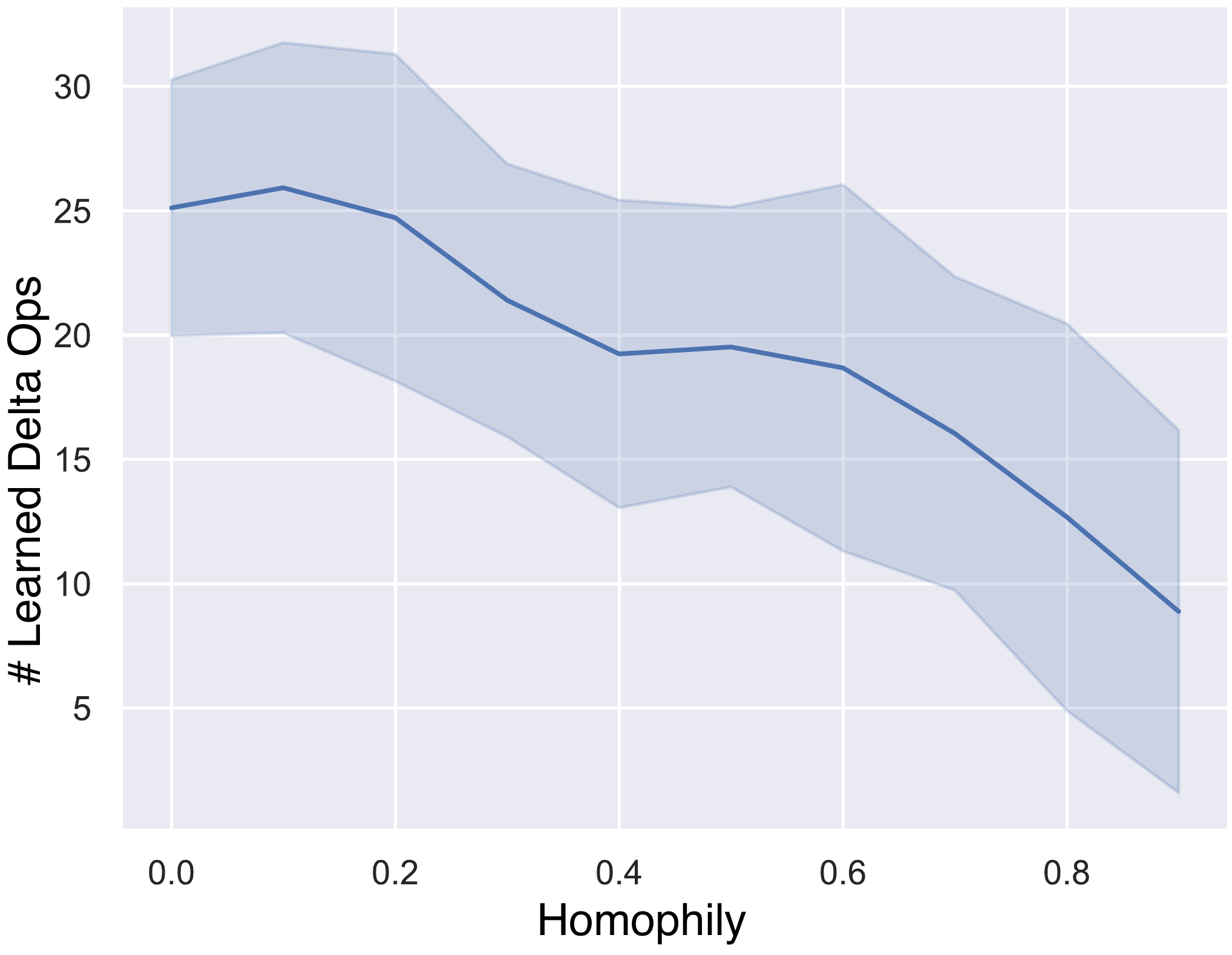}
	\caption{Amount of model capacity devoted to learning delta operators at different levels of homophily.}
	\label{fig:learnedops}
\end{figure}

For each level of homophily, we measured the number of delta operators learned by our model. We present these metrics in Figure \ref{fig:learnedops}. We observe that for low levels of homophily, our model uses ~2.5X of its model capacity on learning delta operators compared with higher homophily. This follows intuition: as the nodes cluster around like-labeled neighbors, the need to identify meaningful feature differences between neighbors at different distances drops significantly. These results strongly suggest that the learned delta operators play a role in the success of \ourmodel\ in Figure \ref{fig:synthetic}.  For this experiment, we trained our model over the synthetic datasets under one constraint: input layer weights $W_0^{(j)}$ are shared across all powers $j \in P$. This allows us to examine sub-columns in the following layer $W_1^{(j)}$. Specifically, we count the number of times a feature, coming out of the first layer, is assigned values of opposite signs in $W_1^{(j)}$. We restrict the analysis to only  values of $W_1^{(j)}$ with magnitude larger than the median in the corresponding column.

\begin{figure*}[t!]
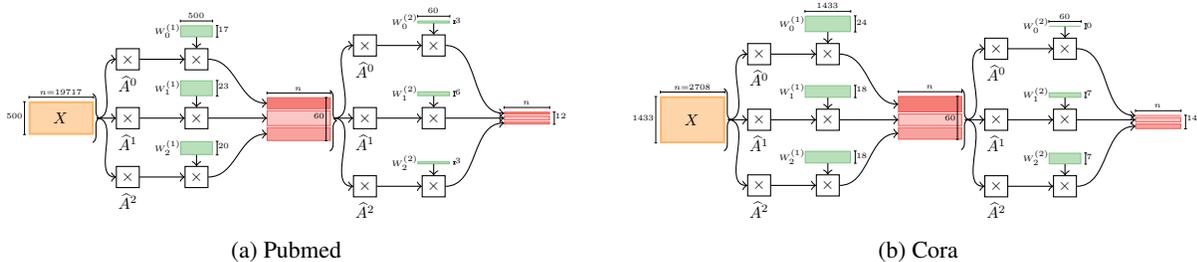

	\centering
	\begin{subfigure}[b]{0.49\textwidth}
		\centering
		\input{figs/learned_pubmed.tex}
		\caption{Pubmed}
	\end{subfigure}%
	\begin{subfigure}[b]{0.49\textwidth}
		\centering
		\input{figs/learned_cora.tex}
		\caption{Cora}
	\end{subfigure}
	\vspace{-0.4cm}
	\caption{
		Learned MixHop Architectures.  Note how different parameter sizes (green boxes) are learned for the two datasets. For example, Group-Lasso regularization on Cora removes all capacity for the zeroth power in the second GC layer. For space, all matrices are plotted transposed and output layer (Section \ref{sec:outputlayer}) has been ommitted.}
	\vspace{-0.3cm}
	\label{fig:learnedarchitectures}
\end{figure*}

\subsection{Node Classification Results}

We show two sets of semi-supervised node classfication results using different splits of our datasets. Because these datasets are taken from the real world, they are inherently noisy, and it is unlikely that achieving 100\% classification accuracy is possible even when given a significant amount of labeled training data. Instead, we are interested in the sparse classification task, namely how well our model is able to improve on previous work while being resilient to noise, even with limited information. 

In Table \ref{tab:planetoid_stats}, we demonstrate how our model performs on common splits taken from \citet{planetoid}. 
Accuracy numbers above double-line are copied from \citet{kipf}.
Numbers below the double-line are our methods, with $P=\{1\}$ being equivalent to vanilla GCNs. $\pm$ represents the standard deviation of 50 runs with different random initializations. All \ourmodel\ models are of same capacity.
These splits utilize only 20 labeled nodes per class during training. We achieve a test accuracy of 71.4\%, 81.9\%, and 80.8\% on Citeseer, Cora, and Pubmed respectively.
Interestingly, for Citeseer, we see that the learned architecture was equal to the original architecture (and so the models performed the same).
In Table \ref{tab:realistic}, we demonstrate how our model performs using random splits with more training information available. These splits utilize 100 nodes per class during training. We achieve a test accuracy of 77.0\%, 87.2\%, and 83.9\% on Citeseer, Cora, and Pubmed respectively. 

As \ourmodel\ is able to pull in linear combinations of features from farther distances, it can extract meaningful signals in extremely sparse settings. We believe this explains why \ourmodel\ outperforms baseline methods in both sets of dataset splits.
The results of these experiments confirm our hypothesis (H2) that higher order graph convolution methods with neighborhood mixing can outperform existing methods on real datasets.

\subsection{Visualizing Learned Architectures}
\label{sec:expmeta}

Figure \ref{fig:learnedarchitectures} depicts the learned architectures for two of the citation datasets.
We note that each dataset prefers its own architecture. 
For example, Cora prefers to have zero-capacity on the $0$th power of the adjacency matrix (effectively ignoring the features of each node) in the second layer. 
Not shown (for space reasons) is Citeseer, which prefers the default parameter settings with the same weight capacity across all powers.
All three real datasets had different final architectures, which confirms our hypothesis (H3) that different architectures are optimal for different graph datasets.

\section{Related Work}
\citet{watchyourstep} uses adjacency powers but for embedding learning. \citet{ngcn, dcnn} use adjacency powers for feature propagation on graphs, but they combine the powers at the end of the network (right before classification), and \citet{highordergraphconv} combine them at the input.
We intermix information from the powers layer-wise, enabling our method to learn neighborhood mixing e.g. delta operators, which contrast the features of immediate neighbors from those further away.
\citet{fast-spectrals} uses more Chebyshev polynomials (i.e. higher-rank) Graph Convolution, but their model underperforms our baseline \citep{kipf}, allowing us to hypothesize that message passing along edges outperforms explicit alignment onto the graph Fourier Basis.

\section{Conclusion}

In this work, we analyzed the expressive power of popular methods for semi-supervised learning with Graph Neural Networks and
we showed they cannot learn general neighborhood mixing functions.
To address this, we have proposed a graph convolutional layer that utilizes multiple powers of the adjacency matrix. 
Repeated application of this layer allows a model to learn general mixing of neighborhood information, including averaging and delta operators in the feature space, without additional memory or computational complexity.
Utilizing L2 group lasso regularization on these stacked layers allows us to learn a unique architecture that is optimized for each dataset.
Our experimental results showed that higher order graph convolution methods can achieve state of the art performance on several node classification tasks.
Our analysis of the experimental results showed that neighborhood difference operators are especially useful in graphs which do not have high homophily (correlation between  edges and labels).
While we focused this paper on applying our proposal to the most popular models for graph convolution, it is possible to implement our method in more sophisticated frameworks including the recent GAT \cite{gat}.
Other recent work like \citep{ying2018hierarchical}, which focuses on hierarchical pooling for  community-aware graph representation  might also be extended to use general neighborhood mixing layers.
\section*{Acknowledgements}
The authors acknowledge support from the Defense Advanced Research Projects Agency (DARPA) under award FA8750-17-C-0106, and acknowledge discussions with
Jesse Dodge and Leto Peel, respectively, on Group Lasso regularization and synethetic experiments.

{\small
\bibliography{highordergc.bib}
\bibliographystyle{icml2019}
}

\end{document}